\documentclass[10pt,twocolumn,letterpaper]{article}

\usepackage{cvpr}
\usepackage{times}
\usepackage{epsfig}
\usepackage{graphicx}
\usepackage{amsmath}
\usepackage{amssymb}

\usepackage[pagebackref=true,breaklinks=true,letterpaper=true,colorlinks,bookmarks=false]{hyperref}

\cvprfinalcopy 

\def\cvprPaperID{6149} 

\begin{document}

\title{Action2Vec: A Crossmodal Embedding Approach to Action Learning}

\author{Meera Hahn, Andrew Silva and James M. Rehg\\
Georgia Institute of Technology\\
Atlanta, GA\\
{\tt\small \{meerahahn, andrew.silva, rehg\} @gatech.edu}
}

\maketitle

\begin{abstract}
We describe a novel cross-modal embedding space for actions, named \emph{Action2Vec}, which combines linguistic cues from class labels with spatio-temporal features derived from video clips. Our approach uses a hierarchical recurrent network to capture the temporal structure of video features. We train our embedding using a joint loss that combines classification accuracy with similarity to Word2Vec semantics. We evaluate Action2Vec by performing zero shot action recognition and obtain state of the art results on three standard datasets. In addition, we present two novel analogy tests which quantify the extent to which our joint embedding captures distributional semantics. This is the first joint embedding space to combine verbs and action videos, and the first to be thoroughly evaluated with respect to its distributional semantics.
\end{abstract}

\section{Introduction}
Many core problems in AI hinge on the successful fusion of language and vision. The past 15 years have seen substantial progress in connecting \emph{nouns to images}, in tasks ranging from image captioning \cite{anderson2017bottom,vinyals2015show,you2016image} to VQA \cite{antol2015vqa} and 
the construction of \emph{joint embedding spaces} for images and text \cite{chen2018text2shape,kiros2014unifying,salvador2017learning}. In contrast, progress in linking text to \emph{videos} has been more limited. One issue is that the semantics of video are defined in terms of actions (verbs), and modern architectures for action recognition treat actions as atomic units, modeling them as discrete labels and ignoring the similarities between action classes. There are good reasons for this approach, as it leads to state-of-the-art classification performance and efficient implementations. However, because there is no commonly agreed-upon taxonomy for actions, different datasets define completely different action categories, which makes it challenging to combine action datasets and reuse models. In fact, action recognition models are highly-specialized to the types of videos and action classes of the dataset they were trained on and cannot be easily repurposed for other tasks. This makes it more difficult to use action recognition for downstream tasks in robotics and AI, because models must always be retrained for the task at hand. In contrast, modern NLP systems routinely combine datasets and repurpose language models for new tasks by leveraging \emph{distributed} word representations such as word2vec. By embedding words into a \emph{continuously-valued} vector space, it is easy to to adapt models to different data domains and corpora. Recently, these representations have been leveraged in constructing joint embeddings for images and text \cite{chen2018text2shape,kiros2014unifying,salvador2017learning}.

They key challenge in linking language and video is the difficulty of building a distributed representation that is suitable for actions. There are two parts to this challenge. The first is the fact that in video the primary lexical unit of meaning is a verb, which defines existence and change. Verbs don't map to regions of pixels in video in the same straight forward way that nouns map to bounding boxes in images. A verb defines a sentence in the same way that an action defines a video: by creating a structure that constrains all other elements. Also while actions often map to visual movement, in real-world videos there are many sources of movement (camera motion, background objects, etc.) which are not part of an action, and in fact some actions can be defined by a relative lack of movement (e.g. sitting and reading). This makes it challenging to construct a mapping from a continuous action space to the space of time-varying pixels in a video. The second source of difficulty stems from the need to evaluate embeddings for both actions and verbs. Note that the quality of a word representation is evaluated via an analogy test~\cite{mikolov2013efficient,mikolov2013distributed}, which uses vector arithmetic in the distributed representation space to test the semantic consistency of the representation. It is a more difficult to create a large number of these tests for verbs than for nouns. Therefore, there are only a small set of analogy tests available for verb embeddings \cite{mikolov2013efficient, mikolov2013distributed} and those that do exist, show that verb embeddings don't carry the same amount of semantic richness that noun embeddings do. This is most likely due to the way that context words are chosen during the embedding process for verbs~\cite{levy2014dependency}. 
In this work, we create new evaluation methods that could be used for any part of speech.

We hypothesize that a \emph{joint embedding space} constructed from video and text is the most attractive way to obtain an effective distributed action representation. Lexical cues define the purpose and meaning of an action and information about how the action is performed is only available from video, and actions that are similar can be expected to have similar video motion patterns. The ability to leverage both types of cues should increase the likelihood of mapping similar semantic action concepts to the same part of the joint embedding space. This paper describes a novel method for constructing a joint embedding space for actions, which we call \emph{Action2Vec}. It is generated by a hierarchical recurrent network model, and which links videos and verbs and provides a novel distributed action representation. Specifically, Action2Vec takes a video clip and associated text (typically a single verb or a verb+noun pair) and maps it to a vector that jointly encodes both the semantic and visual properties of the action. 

We test the effectiveness of Action2Vec in two ways. First, we follow standard practice for image-text embeddings and evaluate the ability of Action2Vec to generalize to novel prediction problems via zero shot learning. Furthermore, our experiment design (see Sec.~\ref{sec:zero-shot-eval}) avoids some problems in addressing domain shift that arose in prior works on zero shot action recognition. We also show that the Action2Vec embedding preserves locality, in the sense that actions that are visually and semantically similar lie close together in the embedding space. We conducted two novel analogy experiments to evaluate the structure of the embedding space and assess their accuracy as a generative representation of actions. First, we use the standard linguistic lexicon WordNet to test the distribution of vectors in our cross-modal (text + video) embedding space and compare it to the Word2Vec embedding space. By comparing different embedding techniques in the form of confusion matrices to WordNet, we are able to test the accuracy and quality of the embeddings for all verbs, something which has not been done before. Second, we evaluate the distributional properties of the embedding space using vector space arithmetic. For example, given two action classes that share the same verb but utilize different nouns, such as ``play piano'' and ``play violin,'' we perform the operation: action2vec(play piano) - word2vec(piano) + word2vec(violin) to yield a novel action descriptor. We show that this descriptor vector is closest to the cross modal embedding for ``play violin.'' Vector arithmetic demonstrates that the multi-modal distributed embedding representation that we have produced retains the semantic regularities of word embeddings. Our results demonstrate that Action2Vec provides a flexible, useful, and interpretable representation for actions in video.
 
This paper makes four contributions. First, we present a method for generating a joint visual semantic embedding of video and text which we refer to as Action2Vec. As part of this work, we will release both our software for learning embeddings and the trained embeddings we generated for all of the benchmark action recognition datasets. Second, we demonstrate the quality of the resulting embeddings by obtaining state-of-the-art results for zero shot recognition of actions. Third, we introduce a new way to test the semantic quality of verb embeddings through the use of confusion matrices. Fourth, we use vector arithmetic to verify the distributional properties of our embeddings and obtain the first vector space results for actions.
\section{Related Work}
\label{relatedwork}

\begin{figure*}[t]
\centering
\includegraphics[scale=.32]{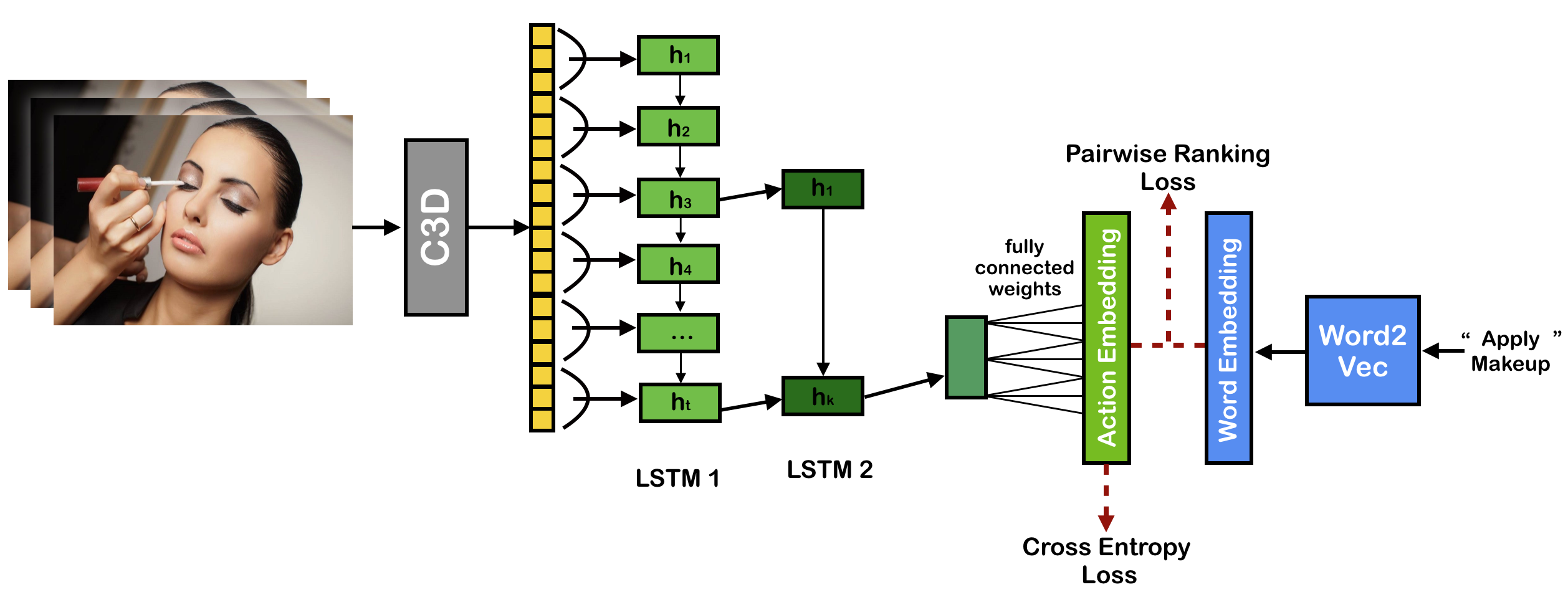}
    \caption{Action2Vec end-to-end architecture for embedding videos into a 300-dimensional vector. Videos are first passed through C3D then and a 2 layer hierarchical LSTM then a fully connected later before going to the two loss functions. Soft attention is added between before the first LSTM and between the 1st and 2nd LSTM. The figure here shows LSTM 1 having filter length of 3 and taking visual feature sub-sequences of 3 as input.}
    \label{fig:network_fig}
\end{figure*}

\noindent\textbf{Distributed Representations:} Our work is based on the now-classic methods for constructing distributional semantic models of language in the form of word embeddings~\cite{mikolov2013efficient,mikolov2013distributed,pennington2014glove}. Mikolov et al. \cite{mikolov2013efficient} is a representative and widely-used example. They introduce a skip-gram-based model to map words into a low-dimensional dense descriptor vector. They demonstrate the embedding's ability to perform analogical reasoning and support compositionality. Our work can be seen as an extension of this approach to leverage video features in constructing a joint embedding. However, our learning architecture is substantially different from~\cite{mikolov2013efficient} and all related prior works, due to the unique aspects of learning from video. The creation of distributed language models led in turn to work on joint image-word and image-sentence embedding spaces~\cite{chen2018text2shape,dong2016word2visualvec,Wang_2016_CVPR,Karpathy_2015_CVPR,kiros2014unifying,salvador2017learning,vinyals2015show}. Of these efforts, the paper by Kiros et. al.~\cite{kiros2014unifying} is perhaps the closest to our work, in that they demonstrate a joint image-text embedding space that supports vector arithmetic. However, our architecture and training approach differ significantly from~\cite{kiros2014unifying}, due to the fact that we are constructing representations from videos instead of still images. A goal of our work is to explore trade-offs between representations that are purely discriminative, optimizing for classification accuracy, and representations that capture the semantic structure of the verb space. We achieve this by optimizing a dual loss that combines a classification loss with a cosine loss Figure~\ref{fig:network_fig}). Other works which have utilized such a dual loss include~\cite{chen2018text2shape,salvador2017learning}.

\noindent\textbf{Video Captioning:} One domain where joint models of video and language arise naturally is in the context of video captioning~\cite{pan2016hierarchical,pan2016jointly}. The task of video captioning faces similar challenges to our problem, but however in captioning the focus is not building and testing a distributed representation but rather focus on the mapping from video to stream of text. Video captioning methods require strong encoder and decoder networks. In our evaluations we demonstrate the effectiveness of our HRNN architecture as an encoder that could possibly be used for captioning models. 

\noindent\textbf{Zero-Shot Action Learning:} Zero-shot learning is a canonical task for evaluating the effectiveness of an embedding space, and several works have used word embeddings to tackle the problem of zero-shot action recognition. The task of zero-shot recognition is to predict the category label for a novel action, which was not known at training time, by mapping it via a previously-trained semantic space~\cite{xu2015semantic,xu2017transductive,xu2016multi,kodirov2015unsupervised,lampert2014attribute,gan2016learning}. Most of the newer works that use deep visual features focus on the domain shift problem. The domain shift occurs when switching from seen to unseen data. In the context of zero shot learning, the unseen test classes are often poorly-explained by the regression mapping that is learned from the training distribution, leading to poor test performance. There are many efforts to ameliorate this problem, however most require the use of auxiliary datasets or self-training. An example is~\cite{xu2015semantic, xu2017transductive}, which requires access to the knowledge of which classes are in the testing set, which runs contrary to our definition of zero-shot learning. Other works use unsupervised approaches \cite{kodirov2015unsupervised} but these methods don't achieve as good of success as ours. In contrast, we approach the domain shift problem through regularization using the unlabeled testing set.

The final body of related work is recent deep learning approaches that construct video representations for supervised prediction tasks such as action recognition. We build on these approaches in our own work. In particular, our model utilizes the C3D \cite{tran2015learning} architecture to extract features from video frames. We also experimented with two-stream approaches similar to \cite{carreira2017quo,feichtenhofer2016convolutional,simonyan2014two}, although in our application we found only minimal benefit from the additional network structure.
\section{Methods}
\label{sec:approach}

\subsection{Problem Formulation}
Our goal is to develop a vector descriptor for every action that captures both the semantic structure of the space of verbs and the spatio-temporal properties that characterize an action in video. We believe we are the first to construct an action embedding that marries features of verbs and video, and we argue that this construction is the key to developing a comprehensive action model which supports both reasoning and classification tasks. Reasoning refers to the ability to make analogies and perform compositions, and otherwise explore the "what" of an action (e.g. what it is for, what are the situations in which it is used, etc.) In contrast, the spatio-temporal pattern of an action captures the "how" of an action (e.g. how is the action performed, how does it transform the world, etc.) and supports classification. We propose an end-to-end method called Action2Vec which obtains an action encoding by using linguistic information from action verb names and combining it with visual information derived from deep video features. To obtain the linguistic information for each action class name, the use the commonly used word embeddings created by the Word2Vec skip-gram model~\cite{mikolov2013efficient}. 

\subsection{Action2Vec Architecture}
The Action2Vec architecture is composed of a Hierarchical Recurrent Neural Network (HRNN) with self attention and a dual cosine similarity and classification loss, illustrated in Figure \ref{fig:network_fig}. The Action2Vec pipeline takes in a video $V=(x_1, x_2, ..., x_n)$ where $n$ is the number of frames in the $V$. We use a HRNN because it provides flexibility in modeling temporal data with greater efficiency then stacked LSTMs. We were motivated by the comparison in~\cite{pan2016hierarchical} between the temporal operations of HRNN and CNN analysis of images. The input to the LSTM 1 is a subsequence of visual features. Every $s$th hidden unit of LSTM 1, where s is the stride of LSTM 1, outputs a vector which becomes the input to LSTM 2. In Action2Vec LSTM 1 acts as local temporal filter over the subsequence of visual features and LSTM 2 aims to learn the dependencies between all the subsequences of the video. The final piece of the HRNN is a fully connected layer transforming the output of the second LSTM to be the same size as the word embedding vectors. We choose to use a weighted dual loss ${L}_{Dual}$ consisting of cross entropy loss ${L}_{CE}$ and pairwise ranking loss ${L}_{PR}$. We do this in order both obtain the information contained in the word vectors whilst retaining the visual information that is used to distinguish between the actions during classification. 

\begin{equation*}
\begin{split}
\label{eqn:pr}
\mathcal{L}_{PR} = \underset{\theta}{\mathrm{min}} \sum_{i} \sum_{x} (1 - s(a_i, v_i)) +  \\ max \{ 0,  s(a_x, v_i) \} + 
 max \{ 0,  s(a_i, v_x) \}
\end{split}
\end{equation*}
\vspace{-.3cm}

\begin{equation*}
\label{eqn:loss}
\mathcal{L}_{Dual} = \mathcal{L}_{PR} + \lambda * \mathcal{L}_{CE}
\end{equation*}

Where $\theta$ represents all the parameters of the network, $a_i$ is an action-video embedding of class $i$, $v_i$ is a verb embedding of class $i$, $a_x$ is an action-video embedding of contrastive class $k$, and $v_x$ is a contrastive verb embedding of class $k$. $s$ is cosine similarity, and $\lambda$ is set to 0.02 after 75\% of the iterations of the first epoch. We did a hyper-parameter search to find the best 
$\lambda$ value and found that a low $\lambda$ value was best at managing the competing losses.

We recognize that in sequencing tasks such as videos temporal relations are especially important. Therefore we use the soft attention mechanism from \cite{bahdanau2014neural}. We add attention units between the visual features and LSTM 1 and between LSTM 1 and LSTM 2. 
Instead of just putting in the sequence of inputs of $(x_1, x_2, ..., x_n)$ to the LSTMS we first apply soft attention to create a new sequence $(c_1, c_2, ..., c_n)$. The attention weights $\alpha$ are calculated at each step t=1,...,n. The attention weight $a^{(t)}_j$ for each $x_i$ is as follows:

\begin{equation*}
\alpha^{(t)}_j = \frac{\exp(e^{(t)}_j)}{\sum^n_{j=1} \exp(e^{(t)}_j)}\\
\end{equation*}

\begin{equation*}
{e}^{(t)}_j=w^\top\tanh(W_a\text{x}_i + U_a\text{\bf{h}}_{t-1} + b_a)
\end{equation*}

Where, ${e}^{(t)}_j$ scores how well the inputs around position $j$ and the output at position $t$ match, also known as the alignment score. $w,W_a,U_a, b_a$ are parameters and $h_{t-1}$ is the hidden state.

For visual features we use the C3D architecture pre-trained on Sports-1M dataset~\cite{KarpathyCVPR14}, given its success in action recognition ~\cite{tran2015learning,pan2016jointly}. We extract C3D features every 16 frames and then use PCA to reduce the dimension to 500. To handle the varying length of videos, we set the all videos to the mean video length of 14 seconds at 30fps. We clipped videos that were longer than 14 seconds and we padded the features-vector sequences with zeros for videos under 14 seconds. Every video is then represented by 52 C3D feature vectors, each with 500 dimensions. First, we divide the 52 C3D vectors into non-overlapping sub-sequences of length 6. For illustration purposes, Figure \ref{fig:network_fig} shows a 18 C3D vectors and sub-sequences length of 3 inputed to the LSTM 1. Each sub-sequence is then passed LSTM 1 for which all units are set to 1024 and dropout of 0.5, which outputs a 1024-dimensional vector for each input sequence. This first LSTM is exploring the local structure of the video sub-sequences. LSTM 1 is shared between all inputs, so the weights do not change depending on which input is being processed, and the hidden states do not carryover between forward passes. We set LSTM 1 to have a filter length of 8 which means that for every eighth hidden unit, it produces a 1x1024 vector. For illustration purposes Figure \ref{fig:network_fig} shows a filter length of 3. We found through experimentation this filter length to keep quality embeddings without increasing computation. The output of LSTM 1 is then passed into a second LSTM, for which all units are set to 512 and no dropout and only outputs a single 512-dimensional vector. This output then passes through a 300 dimensional fully-connected layer to transform the output to the same dimension as the word embedding labels.

The output of the HRNN is directly compared against the word embeddings using a pairwise ranking loss ${L}_{PR}$. Using the cross-entropy classification loss ${L}_{CE}$, the 300-dimensional vector goes through one final fully-connected layer with the same dimensionality of the one-hot action class labels, and with a sigmoid activation. The entire network is trained the result of the dual loss ${L}_{Dual}$. Since we are training with a pairwise ranking loss we use hard negative mining. To handle the competing losses we had to scale ${L}_{CE}$ to be smaller using $\lambda$. The network is optimized with Adam~\cite{kingma2014adam}. 

We use the word vectors of class names as labels for the pairwise-ranking loss. Certain class names are not in verb form, so we edit the names to an equivalent word that exists in the Word2Vec model. For example the class name ``walking'' is a adjective and noun, so it is changed to the verb ``walk''. Similarly, the class name ``clean and jerk'' was not in the Word2Vec model, so the name was changed to the analogous name of ``weightlift.'' For class names that are longer than a single word, we average the word vectors that make up the name. 

\section{Evaluation}
\label{sec:evaluation}
In this section, we present the results from extensive evaluation of the Action2Vec embeddings. First, in \S\ref{sec:zero-shot-eval} we validate the quality of our network's ability to project new videos into the joint space by performing zero shot action recognition and obtaining state of the art results. Second, in \S\ref{sec:analogytests} we present \emph{two novel forms of analogy tests} to validate the semantics of our representation. In the first test, we construct confusion matrices using the similarities between verbs based on Action2Vec and Word2Vec, and then compare the confusion matrices using ranking correlation. This novel approach allows us to directly rate the distances for all verbs in the vocabulary, something that has not been done previously. Our second test performs a systematic and thorough evaluation of vector arithmetic for action embeddings, a topic which has not been explored in previous works. Even within the NLP literature, the standard testing sets for verbs are limited to verb tenses~\cite{agirre2009study,gladkova2016analogy}. Collectively, these experiments constitute the most thorough evaluation of verb embeddings that has ever been performed. 

\paragraph{Datasets:}
Our evaluations are based on three standard datasets for action recognition: UCF101 \cite{soomro2012ucf101}, HMDB51 \cite{kuehne2013hmdb51} and Kinetics \cite{kay2017kinetics}. We selected these datasets because they contain diverse videos for each action class, allowing us to test the generalization of our method to actions in various environments, poses, and contexts. For example, in the HMDB51 dataset, the action class ``push'' has a variety of videos, from children pushing toy trains to adults pushing tables. All datasets are focused on human activities. This paper is the first to use Kinetics for zero-shot learning. We use Kinetics to show that our embeddings can scale to larger datasets as well as to obtain the most diverse set of embeddings by using a dataset with a large number classes.

\subsection{Zero-Shot Action Recognition}
\label{sec:zero-shot-eval}
Our experimental set up has a labeled training set and unlabeled testing set. The labels of the training set and testing set have no overlap, and the labels of the testing set are never seen by the Action2Vec model before test time. We train each ZSL method on the videos and verb embeddings of our training set. We then use the trained model to encode all videos in the test dataset. These predicted video vectors are then normalized and assigned the label of the nearest neighbor verb embedding. The nearest neighbor is calculated using cosine distance. The accuracy is calculated by the percentage of the test videos for which a model is able to correctly predict the action class. We test on the all datasets described above. For each dataset, we test on 3 different amounts of held out action classes: 50\%, 20\% and 10\%. We observe that performance decreases as we withhold a greater number of classes. This is expected since the model has less information with which to interpret new action classes. We recognize that even when only 10\% of classes are held out for testing, we do not achieve accuracy similar to that for classical action recognition. We attribute gap in performances to the open problem of overcoming domain shift which happens when switching to an entirely new set of prediction classes. First we dissect our network with an ablation study over the three train-test splits in Table \ref{zeroshot_Ablation}. We show the the comparison of the best Action2Vec architecture with all relevant prior work in Table \ref{zeroshot_comparison}. 

\begin{table*}[!htp]
    \begin{center}
    \tabcolsep=0.4cm
    \begin{tabular}{llll}
    ~                                                        & HMDB51 & UCF101 & Kinetics \\ \hline
    \textbf{50/50}                   & ~      & ~      & ~        \\ \hline
    Action2Vec w/dual loss + attention                   & \textbf{23.48}  & \textbf{22.10}  & \textbf{17.64}        \\
    Action2Vec w/dual loss         &                22.39  & 21.63  & 15.35        \\
    Action2Vec w/only CosSim loss         & 21.34   & 21.11   & 14.62        \\
    Stacked LSTM w/dual loss        & 17.65   & 17.13  & 12.98       \\
    Single LSTM w/dual loss        & 18.09   & 17.86   & 13.40        \\
    Pooled C3D fc7                   & 5.00   & 11.41   & 9.89        \\
    \hline
    \textbf{80/20}                   & ~      & ~      & ~        \\ \hline
    Action2Vec w/dual loss + attention                   & \textbf{40.11}  & \textbf{36.51}  & \textbf{22.93}        \\
    Action2Vec w/dual loss   & 39.85  & 35.82  & 22.34        \\
    Action2Vec w/only CosSim loss         & 38.16   & 36.39   & 21.55        \\
    Stacked LSTM w/dual loss        & 35.23   & 32.94  & 18.27       \\
    Single LSTM w/dual loss       & 34.30   & 32.00   & 15.26        \\
    Pooled C3D fc7                   & 8.77   & 23.89   & 18.76   \\ \hline
    \textbf{90/10}                   & ~      & ~      & ~        \\ \hline
    Action2Vec w/dual loss + attention                   & \textbf{60.24}  & \textbf{48.75}  & \textbf{38.02}        \\
    Action2Vec w/dual loss         & 58.01  & 47.94  & 36.93        \\
    Action2Vec w/only CosSim loss         & 51.85   & 46.44   & 36.88        \\
    Stacked LSTM w/dual loss        & 49.83   & 42.43  & 30.62       \\
    Single LSTM w/dual loss       & 53.76   & 44.74   & 32.31        \\
    Pooled C3D fc7                   & 23.11   & 36.29   & 26.31        \\
    \end{tabular}
    \end{center}
   \caption{Ablation study of the Action2Vec architecture: ZSL classification accuracy. Shows the accuracy for 3 types of train-test splits. In example, 80:20 means the model was trained on 80\% of the dataset classes and 20\% of the classes were only seen during test time.} \label{zeroshot_Ablation}
\end{table*}

\begin{table*}[!htp]
    \begin{center}
    \begin{tabular}{llll}
    Train-Test Split                               & HMDB51 & UCF101 & Kinetics \\ \hline
    \textbf{50/50}                   & ~      & ~      & ~        \\ \hline
   Action2Vec w/dual loss + attention  & \textbf{23.48}  & \textbf{22.10}  & \textbf{17.64} \\
    Pooled C3D fc7  & 5.00  & 11.41  & 9.89 \\
    TZS w/out aux data + w/ knowledge test labels \cite{xu2017transductive}
    & 19.10  & 20.8  & - \\
    TZS w/out aux data + w/out knowledge test labels \cite{xu2017transductive}
    & 14.50  & 11.70  & - \\
    SAV w/out aux data \cite{xu2016multi}  & 15.00  & 15.80  & - \\
    UDA \cite{kodirov2015unsupervised} & -  & 14.00  & - \\\hline
    \textbf{80/20}                   & ~      & ~      & ~        \\ \hline
   Action2Vec w/dual loss + attention  & \textbf{40.11}  & \textbf{36.51}  & \textbf{22.93} \\
    KDCIA \cite{gan2016learning}  & -  & 31.1  & -\\
    KDCIA \cite{gan2016learning}  & -  & 29.6  & -\\
    UDA \cite{kodirov2015unsupervised}  & -  & 22.50  & -\\

    \end{tabular}
    \end{center}
   \caption{Comparison of accuracy for different ZSL methods.} \label{zeroshot_comparison}
\end{table*}
\paragraph{Zero-Shot Action2Vec Ablation:}  The ablation study of Action2Vec shows that our network is superior to other possible architectures and justifies our use of a dual loss and the soft attention mechanism. We show that the HRNN achieves better accuracy than a 2-layer stacked LSTM. It does this while also taking less time to train than the stacked LSTM. We find that using both LSTM layers significantly enhances performance over using a single LSTM. We attribute this to the second LSTM accounting for temporal relations between video snippets that the first LSTM could have missed as well as that two LSTMs do a more gradual dimensionality reduction. Action2Vec's better performance on HMDB51 compared to UCF101 is interesting as HMDB51 is smaller and has greater scene diversity. Superior performance on a smaller and harder dataset suggests that Action2Vec is capturing meaningful semantic and temporal properties. Additionally the poor performance of the pooled C3D baseline demonstrates the importance of using a recurrent model to capture and encode the temporal information that may be lost when using only 3D convolutions. 

\paragraph{Zero-Shot Baselines:} The results in Table \ref{zeroshot_comparison} demonstrate that the Action2Vec embeddings outperform all baseline methods in every data split, for every dataset. The baselines methods \cite{xu2017transductive} and \cite{xu2015semantic} use low-level features such as HOG to create the video embeddings and then trains a Kernel Ridge Regression model to map the action space to the semantic space. We report results of their architecture when not using any auxiliary training data. TZS requires access to the (transductive) access to test data which slightly pushes the bounds of our ZSL experimental setup so we report their results both with and without transductive access to the test data. The baseline \cite{gan2016learning} uses attribute detection to do ZSL. The baseline \cite{kodirov2015unsupervised} takes a unsupervised approach to addressing domain shift. Additionally we show show results when using pooled C3D features ~\cite{pan2016jointly} to construct the video embeddings. We train a Kernel Ridge Regression model with Laplacian regularization to map the pooled C3D vectors to the word vector labels. Despite other methods using deep features, our architecture out preforms all others. 

\subsection{Analogy Tests}
\label{sec:analogytests}
In natural language processing, representations of distributional semantics are commonly evaluated using analogy tests, which take the form: $vector_1 - vector_2 + vector_3 = vector_4$. The test is passed if the word corresponding to $vector_4$ makes logical sense. For example, in the analogy $King - Man + Woman$, the resulting vector should represent the word $Queen$. Analogy tests are the gold standard for evaluating distributional semantics because they assess the relational capacity of the vector space. There are a few standard manually constructed analogy test sets~\cite{mikolov2013efficient,gladkova2016analogy,agirre2009study}. Unfortunately, these test sets are not comprehensive, covering nouns thoroughly but not adequately testing verbs. In fact, there are no existing test sets whose verb coverage is sufficient, as current evaluations only test the ability to translate between verb tenses. An example of an standard test is ``accept is to acceptable as achieve is to what,'' with the answer being ``achievable''~\cite{gladkova2016analogy}. To address this issue, we \emph{introduce two new methodologies} for conducting verb analogy tests which can scale to any dataset. 

\begin{figure}[!htbp]
\begin{center}
\includegraphics[scale=.25]{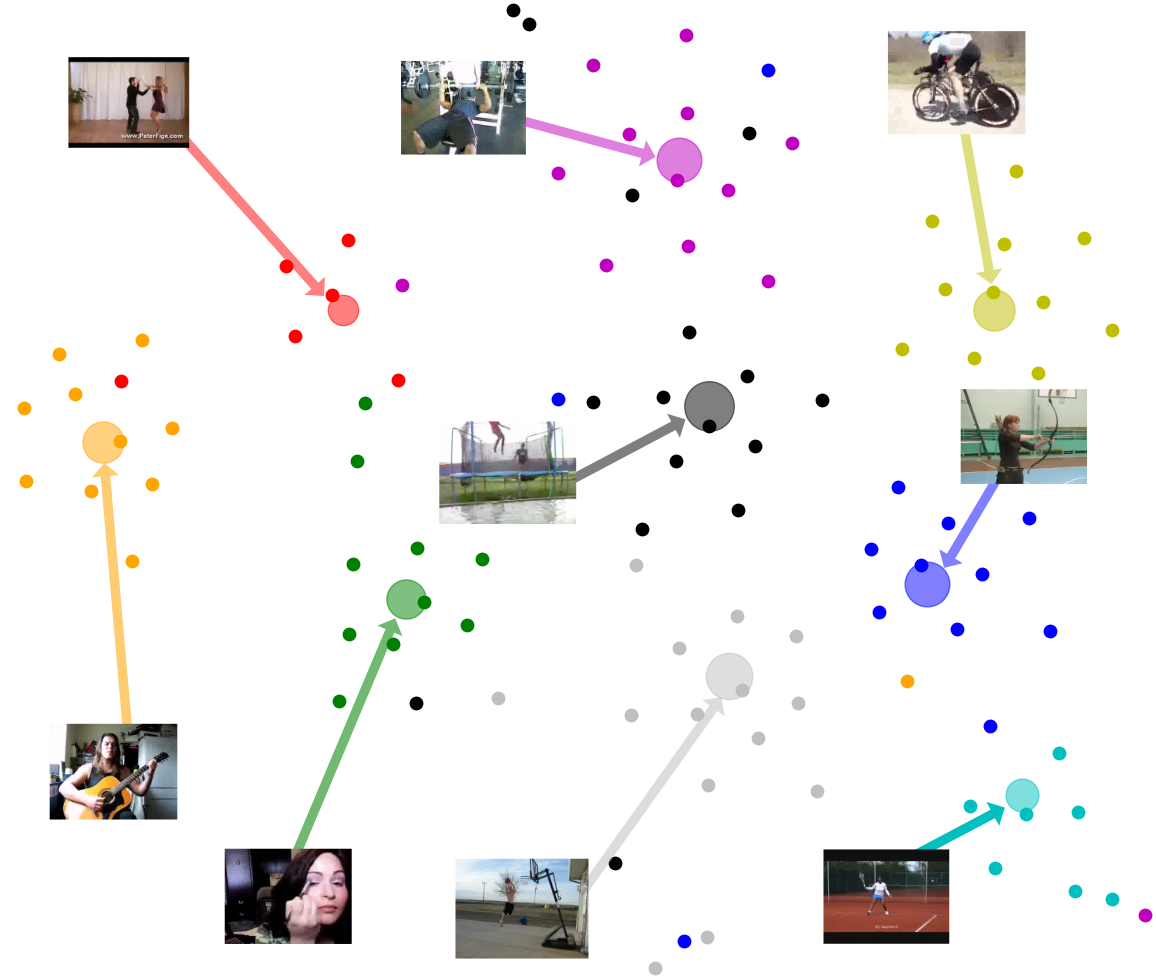}
\end{center}
   \caption{Tsne visualization of the Action2Vec embeddings of the action classes in UCF101.}
\label{fig:long}
\label{fig:onecol}
\end{figure}

\subsubsection{Word Matrices}
WordNet is a large and popular English lexical database that contains the majority of words in the English language~\cite{wordnet}. The words in this database are grouped into synsets, which form a graph of the lexical and semantic relations between the words. Using this graph we can measure the relations between individual words. Specifically, the Wu-Palmer algorithm can be used to measure the semantic similarity between any two words in WordNet~\cite{wu1994verbs}. The relational distances between words in WordNet can be seen as ground truth measurements since WordNet was manually constructed by linguists. 

\begin{table}[!htbp]
    \begin{center}
    \tabcolsep=0.11cm
    \begin{tabular}{|lccc|}
    \hline
    ~     & HMDB51 & UCF101 & Kinetics \\ \hline
    \multicolumn{1}{|l|}{WordNet vs. PooledC3D}  & 0.0716  & 0.0773  & 0.0929 \\
    \multicolumn{1}{|l|}{WordNet vs.  Word2Vec}  & 02855  & 0.2292  & 0.2807  \\
     \multicolumn{1}{|l|}{WordNet vs. Action2Vec } & 0.2353  & 0.2092  & 0.2421   \\ \hline
    \end{tabular}
    \end{center}
   \caption{Shows the Spearman Rank Correlation between the gold standard WordNet similarity confusion matrix and the confusion matrices created for the 3 types of verb embeddings.} 
\label{rankCorrelation}
\end{table}

We now describe our novel construction of word similarity matrices and their use in evaluating our embedding. For each dataset, we take the list of action class names and remove all names that are not in WordNet. Then for each dataset we create a confusion matrix for the list of classes, where the values of the matrix at a given index corresponded to the Wu-Palmer WordNet similarity distance~\cite{wu1994verbs}. Then we take the corresponding Action2Vec embeddings for the class name list and create the same confusion matrix, except that now the values of the matrices at a given index corresponded to the cosine similarity between the two Action2Vec embeddings. In order to get a single embedding for an action class, we average the Action2Vec vectors for all videos of that action class. Finally, we create a confusion matrix using the corresponding Word2Vec embeddings and another using the pooled C3D embeddings. 

Since we are taking the Wu-Palmer WordNet measurement to be the ground truth similarity confusion matrix, we compare it individually to the Action2Vec, Word2Vec, and C3D similarity confusion matrices. Since the measurements of cosine similarity and Wu-Palmer similarity are at different scales, we compare the matrices using Spearman Ranking Correlation~\cite{kendall1955rank}. 
We can only calculate ranking correlation between pairs of rows, so we average the ranking correlations across all rows and use that as the ranking correlation between two matrices. Ranking correlation lies between -1 and 1, with 1 being the best possible score and identical ranks. The results of the ranking correlations between all matrices are shown in Table \ref{rankCorrelation}.

From Table \ref{rankCorrelation} reveals that the Word2Vec matrix has the highest correlation with WordNet in every dataset. This is to be expected, as Word2Vec has been trained specifically for the task of linguistic representation. In contrast, the pooled C3D vectors are significantly worse than the other two embeddings. This demonstrates that the relational semantic information of a purely visual encoding of a video is relatively shallow. Action2Vec matrix only gets a slightly lower rank correlation than the Word2Vec matrix. This shows us that the Action2Vec architecture, in addition to capturing visual information, is apt at semantically encoding the similarities between actions.

\subsubsection{Vector Arithmetic}
Here we describe a \emph{novel analogy test for actions} based on vector arithmetic, following previous analogy tests in text-image embeddings~\cite{kiros2014unifying}. This tests the robustness of our action representations by testing how they generalize over different noun combinations. This experiment is designed to easily scale to most action datasets, in which multiple action classes contain the same verb with different nouns. For example, in Kinetics the verb ``throw'' has 8 instances ranging from throwing an ax to throwing a Frisbee. We perform the following vector arithmetic for each verb with multiple noun instances: $verb \ noun1 - noun1 + noun2 = V_{new}$.
We then search the vector space of action classes and select the class whose vector is closest to $V_{new}$ in euclidean distance. If $V_{new} = verb \ noun2$ we count this as correct. An example: $throw\ softball - softball + football = throw\ football$. Just as in the word matrix analogy tests, in order to get a single embedding for an action class, we average the Action2Vec vectors for all videos of that action class. We performed this test for Kinetics and UCF101 but not HMDB51, as it is too small and does not have multiple nouns per class. From Table \ref{vectorarthemtic} we can see that the vector arithmetic does not perform as well on Kinetics as it does on UCF101. We found that it most commonly fails on ambiguous actions such as ``doing'' and ``making,'' which are missing from UCF101, making Kinetics more difficult. We were unable to preform this experiment on the HMDB51 dataset because it does not contain multiple action classes with the same verb. 

\begin{table}[!htbp]
    \begin{center}
    \tabcolsep=0.35cm
    \begin{tabular}{|lcc|}
    \hline
    ~     & UCF101 & Kinetics \\ \hline
    \multicolumn{1}{|l|}{Number of Comparisons}  & 90  & 1540   \\
    \multicolumn{1}{|l|}{Average Precision}  & 0.9875  & 0.5755     \\
   \hline
    \end{tabular}
    \end{center}
   \caption{Vector arithmetic analogy test results on UCF101 and Kinetics. First row: total number of analogy tests for each dataset. Second row: the percentage of tests that passed.} \label{vectorarthemtic}
\end{table}

\begin{figure*}[!htbp]
\centering
\includegraphics[scale=.25]{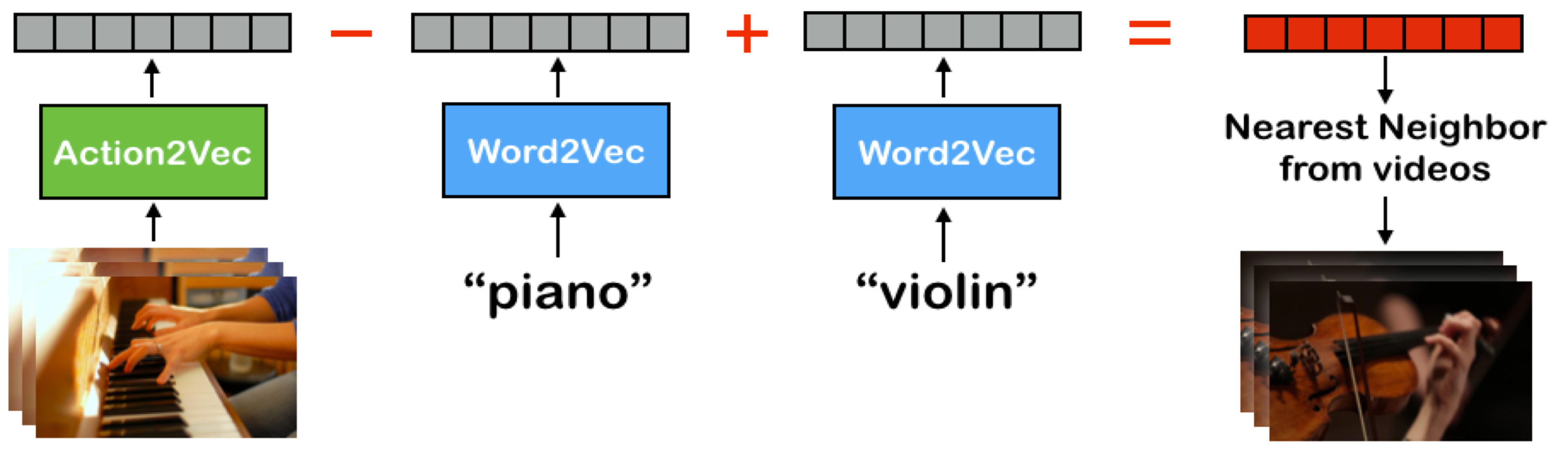}
    \caption{Vector Arithmetic Analogy Test: For a given instance we preform the following vector manipulation. We get the mean Action2Vec embedding for the videos of the first action class 'play piano'. We subtract the Word2Vec embedding of the first noun and add the Word2Vec embedding of the second noun. Then we search the Action2Vec space for nearest neighbor to the resulting vector. The test is passed if the nearest neighbor is the Action2Vec embedding for the second action class, which in this case is 'play violin'.}
    \label{fig:vector_arithmetic}
\end{figure*}

\begin{figure*}[!htbp]
\centering
\includegraphics[scale=.48]{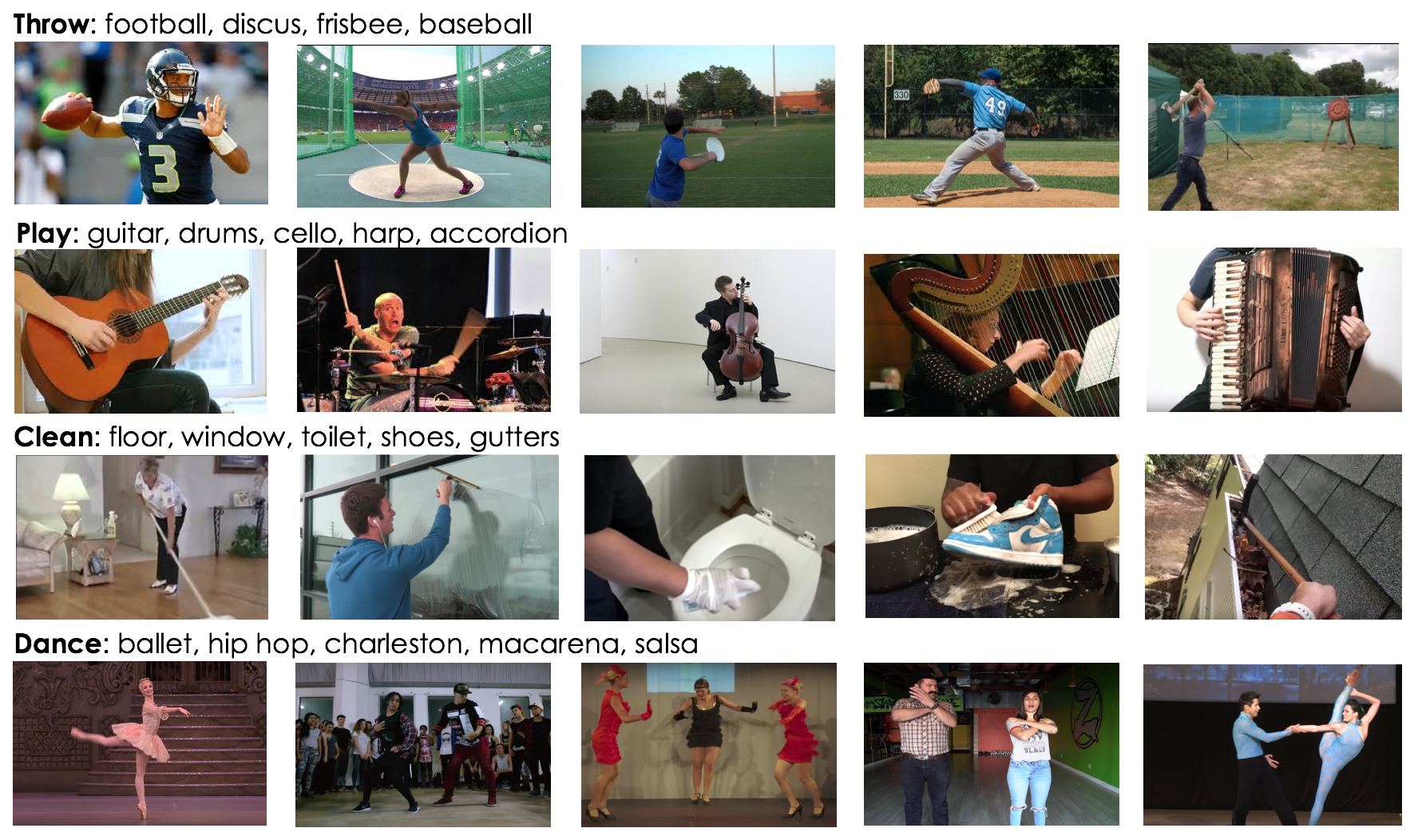}
    \caption{Examples of the vector arithmetic analogy tests.}
    \label{fig:vec_examples}
\end{figure*}
\section{Conclusion}
The integration of vision and language is increasingly important for the advancement of AI. While joint text and image representations have been studied, the combination of text and video remains largely unexplored. Video is a natural modality for leveraging verb semantics, and the relationship between a verb and its video representation is complex. This paper introduces \emph{Action2Vec}, a novel end to end embedding model for actions which combines linguistic cues with spatio-temporal visual cues derived from deep video features. We demonstrate that Action2Vec effectively captures discriminative visual features by delivering state-of-the-art zero shot action recognition performance. Additionally we introduce a new way to test both action and verb embeddings and using this method we show that Action2Vec has also closely captures the linguistic semantics of WordNet. We believe this is the first thorough evaluation of a video-verb embedding space with respect to accuracy and semantics. We hope that Action2Vec can provide a useful intermediate representation for tasks in video generation, video retrieval, and video question answering.
\clearpage

{\small
\bibliographystyle{ieee}
\bibliography{egbib}
}

\end{document}